\documentclass[conference]{IEEEtran}
\IEEEoverridecommandlockouts
\usepackage{cite}
\usepackage{amsmath,amssymb,amsfonts}
\usepackage{graphicx}
\usepackage{subcaption}
\usepackage{textcomp}
\usepackage{xcolor}
\usepackage{algorithm}
\usepackage{algorithmicx}
\usepackage{algpseudocode}
\usepackage{tikz}
\algblock{ParFor}{EndParFor}
\algnewcommand\algorithmicparfor{\textbf{parfor}}
\algnewcommand\algorithmicpardo{\textbf{do}}
\algnewcommand\algorithmicendparfor{\textbf{end\ parfor}}
\algrenewtext{ParFor}[1]{\algorithmicparfor\ #1\ \algorithmicpardo}
\algrenewtext{EndParFor}{\algorithmicendparfor}
\def\BibTeX{{\rm B\kern-.05em{\sc i\kern-.025em b}\kern-.08em
    T\kern-.1667em\lower.7ex\hbox{E}\kern-.125emX}}

\makeatletter
\def\endthebibliography{%
	\def\@noitemerr{\@latex@warning{Empty `thebibliography' environment}}%
	\endlist
}
\makeatother

\newcommand\copyrighttext{%
  \footnotesize \textcopyright 2020 IEEE. Personal use of this material is permitted.
  Permission from IEEE must be obtained for all other uses, in any current or future
  media, including reprinting/republishing this material for advertising or promotional
  purposes, creating new collective works, for resale or redistribution to servers or
  lists, or reuse of any copyrighted component of this work in other works.}
\newcommand\copyrightnotice{%
\begin{tikzpicture}[remember picture,overlay]
\node[anchor=south,yshift=10pt] at (current page.south) {\fbox{\parbox{\dimexpr\textwidth-\fboxsep-\fboxrule\relax}{\copyrighttext}}};
\end{tikzpicture}%
}

\begin{document}

\title{A Novel DNN Training Framework via Data Sampling and Multi-Task Optimization
}

\author{
	\IEEEauthorblockN{Boyu Zhang, A. K. Qin, Hong Pan and Timos Sellis}
	\IEEEauthorblockA{\textit{Department of Computer
Science and Software Engineering} \\
\textit{Swinburne University of Technology}\\
Melbourne, Australia}
    \IEEEauthorblockA{Email: boyuzhang@swin.edu.au, kqin@swin.edu.au, hpan@swin.edu.au, tsellis@swin.edu.au}
}

\maketitle
\copyrightnotice

\begin{abstract}
Conventional DNN training paradigms typically rely on one training set and one validation set, obtained by partitioning an annotated dataset available for the purpose of training, namely gross training set, in a certain way. The training set is used for training the model while the validation set is used to estimate the generalization performance of the trained model as the training proceeds to avoid over-fitting. There exist two major issues in this training paradigm. Firstly, the validation set may hardly guarantee an unbiased estimate of the generalization performance due to potential mismatching with the test data. Secondly, training a DNN corresponds to solve a complex optimization problem, which is prone to getting trapped into inferior local optima and thus leads to the undesired training result. To address these issues, we propose a novel DNN training framework. It generates multiple pairs of training and validation sets from the gross training set via random splitting, trains a DNN model of a pre-specified network structure on each pair while making the useful knowledge (e.g., promising network parameters) obtained from one model training process to be transferred to other model training processes via multi-task optimization (i.e., a recently emerging optimization paradigm), and outputs the best one, among all trained models, which has the overall best performance across the validation sets from all pairs. The knowledge transfer mechanism featured in this new framework can not only enhance training effectiveness by helping the model training process to escape from local optima but also improve on generalization performance via implicit regularization imposed on one model training process from other model training processes. We implement the proposed framework, parallelize the implementation on a GPU cluster, and apply it to train several widely used DNN models. Experimental results on several classification datasets of different nature demonstrate the superiority of the proposed framework over the conventional training paradigm.
\end{abstract}

\begin{IEEEkeywords}
Multi-task optimization, MTO, training deep neural networks, data sampling.
\end{IEEEkeywords}

\section{Introduction}

Deep neural networks (DNNs) have achieved performance breakthroughs in many real-world applications due to their powerful feature learning capabilities, which are typically characterized by sophisticated architectures involving a massive amount of parameters. Training a DNN is equivalent to solving a highly complex non-convex optimization task which is easily stuck into inferior local optima and accordingly leads to the undesired training result.

A commonly employed way to train a DNN relies on an available training set, where a certain loss function defined on the training set is optimized with respect to network parameters to derive optimal parameter values. This optimization process, a.k.a. training process, is iterative and usually terminated by the pre-specified maximal number of training epochs. However, the way of manually specifying the maximum number of training epochs is often too subjective, which increases the risk of over-fitting or under-fitting and accordingly results in undesired generalisation performance. To address this issue, another training paradigm has gained much popularity nowadays, which partitions the original training set, namely the gross training set, into one training set and one validation set via a certain sampling method and utilizes the validation set to estimate the generalization performance of the trained model as the training process (based on the training set) proceeds \cite{prechelt1998early,prechelt1998automatic}. This way may improve the generalization performance of the trained model. However, the validation set may not well represent the potential test data and thus become less effective to provide an unbiased estimate of generalization performance.

To address the above issues, we propose a novel DNN training framework which formulates multiple related training tasks by using a certain sampling method to generate multiple different pairs of training and validation sets from the gross training set and solves these related tasks simultaneously via a newly emerging multi-task optimization (MTO) technique that allows the useful knowledge (e.g., promising network parameters) obtained from one training task to be transferred to other training tasks. Specifically, this framework generates multiple pairs of training and validation sets from the gross training set via a specific sampling method, trains a DNN model of a pre-specified network structure on each pair while enabling the useful knowledge obtained from one training process to be shared with other training processes via MTO, and finally outputs the best one, among all the trained models, which achieves the overall best performance across the validation sets from all pairs. The knowledge transfer and sharing mechanism featured in the proposed framework can not only enhance training effectiveness by helping the training processes to escape from local optima but also improve on generalization via implicit regularization imposed on one training process which comes from other training processes. It is worth noting that the cross-validation technique \cite{arlot2010survey} commonly used to improve generalization performance when training machine learning (ML) models is for tuning the hyper-parameters of the ML model instead of the model parameters per se. Therefore, it is irrelevant to this study which is focused on learning the parameters (i.e., connection weights and biases) of a DNN model with pre-specified hyper-parameters. Another machine learning technique named ensemble learning also trains multiple models to make a prediction. However, it aims at achieving the best performance by assembling all models in a certain way while our proposed framework aims to train a single best model with the help from training other models.

We implement the proposed training framework, parallelize the implementation on a GPU cluster, and apply it for training three popular DNN models, i.e., DenseNet-121 \cite{huang2017densely}, MobileNetV2 \cite{sandler2018mobilenetv2} and SqueezeNet \cite{iandola2016squeezenet}. Performance evaluation and comparison on three classification data sets of different nature demonstrate the superiority of the proposed training framework over the conventional training paradigm in terms of the classification accuracy obtained on the testing set. 

In the following, we will first introduce the background of this work in section \ref{bg}, then describe the proposed framework and its implementations in detail in section \ref{mt}, and finally discuss and analyze experimental results in section \ref{ex}, followed by concluding remarks and future work in section \ref{cl}.

\section{Background\label{bg}}

\subsection{Training Deep Neural Networks}
Gradient descent based optimization algorithms are widely used for training DNNs, e.g., in supervised learning problems. These algorithms normally use back propagation (BP) \cite{leung1991complex} to calculate the gradients of DNN's parameters and accordingly update parameter values. Specifically, given a training set composed of multiple pairs of the input and the output (i.e., the label of the input), a loss function is formulated which measures the mismatch between the output of the network w.r.t. an input and the actual output of that input in the training set summed over all input-and-output pairs in the training set. Then, BP with stochastic gradient descent is applied to calculate the partial derivative of the loss function with respect to each parameter in the DNN from the last layer to the first layer. Next, parameter values are updated based on the calculated derivatives via a certain learning rule. This training process is iterated until a certain stopping criterion is met. 

Training DNNs relies on an available training set which may be used in different ways, leading to different training paradigms. One common practice is to directly train a DNN on the gross training set (i.e. the original training set) and use the pre-specified maximum number of training epochs to terminate the training process. However, the subjective choice of the maximum number of training epochs is likely to make the trained model to overfit the training set and thus lead to inferior generalization. Another popular training paradigm addresses this issue by partitioning the gross training set into one training set and one validation set via random splitting and using the validation set to estimate the generalization performance
of the trained model as the training process proceeds and accordingly stop the training process if the generalization performance of the trained model cannot be much improved \cite{le2011optimization,springenberg2015unsupervised,jaderberg2017population,sabour2017dynamic,hinton2015distilling}.

In ML, cross-validation is a widely used strategy to improve the generalization performance of a trained ML model. However, it is merely applied to tune the hyper-parameters of the model \cite{huynh2005effective}, e.g., the number of layers, the number of hidden neurons, and the learning rate in the context of DNN training. In this work, we focus on learning the parameters (i.e., connection weights and biases) of a DNN model with pre-specified hyper-parameters. Therefore, cross-validation strategy is irrelevant to this study.

\subsection{Multi-Task Optimization}
MTO investigates how to effectively and efficiently tackle multiple optimization tasks concurrently via online knowledge transfer. This paradigm has been inspired by the well-established concepts of transfer learning \cite{pan2009survey} and MTL \cite{caruana1998} in predictive analytics. Existing MTO techniques are mainly developed for Bayesian optimisation \cite{swersky2013multi,bardenet2013collaborative,yogatama2014efficient} or evolutionary computation techniques \cite{gupta2016multifactorial,feng2018evolutionary,zhang2018evolutionary,gupta2016genetic}. Swersky \textit{et al.} \cite{swersky2013multi} proposed multi-task Bayesian optimisation (MTBO) which is based on the well-studied multi-task Gaussian process models. This work can transfer the knowledge gained from prior optimisations to new tasks in order to find optimal hyperparameter settings more efficiently, or optimise multiple tasks simultaneously when the goal is maximizing average performance, e.g., optimizing $k$-fold cross-validation. \cite{gupta2015multifactorial} proposed an MTO based evolutionary algorithm (EA) where tasks benefit from the implicit knowledge transfer during the task solving and can often lead to accelerated convergence for a variety of complex optimization functions. Besides EA, other variants \cite{feng2017empirical,zhong2018multifactorial} are also developed to solve the MTO problems. In \cite{feng2018evolutionary}, Feng \textit{et al.} proposed an evolutionary multitasking algorithm with explicit knowledge transfer via denoising autoencoder, which demonstrates higher efficacy over implicit knowledge transfer. \cite{zhang2018evolutionary} proposed an MTO based framework of generating feature subspaces for ensemble classification.

\section{Proposed Method\label{mt}}
\begin{figure*}
	\centering
	\begin{subfigure}[t]{0.45\linewidth}
		\centering\includegraphics[width=\linewidth]{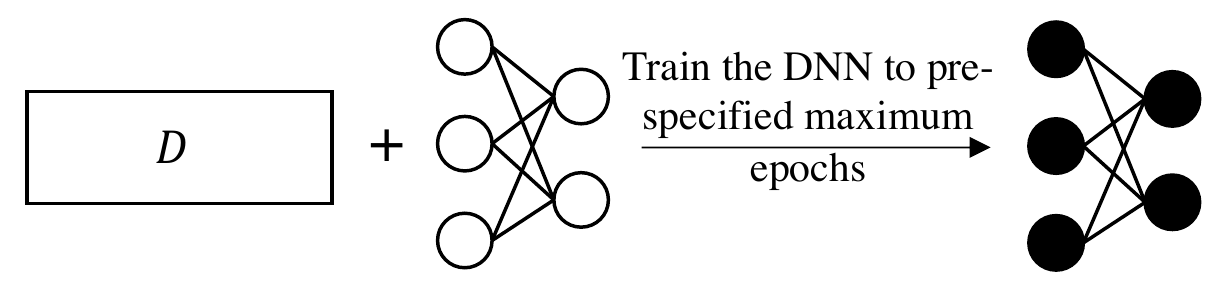}
		\caption{STO: Training the DNN without validation set.}
		\label{stowov}
	\end{subfigure}\\
	\begin{subfigure}[t]{0.7\linewidth}
		\centering\includegraphics[width=\linewidth]{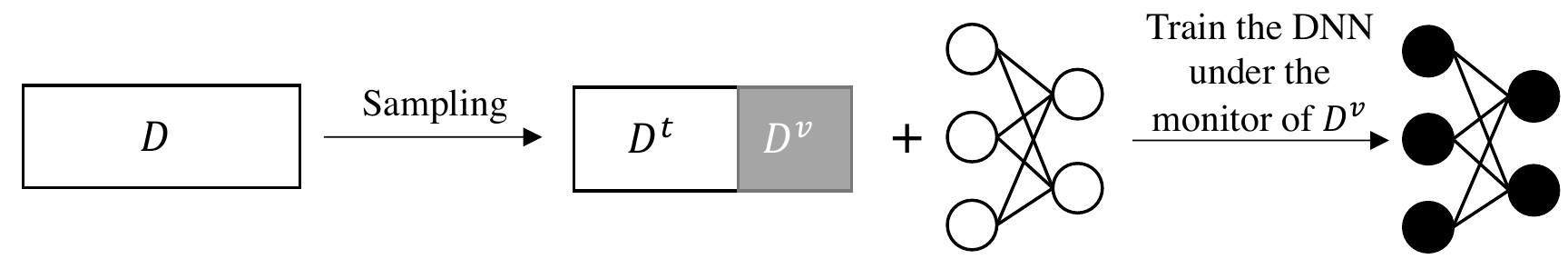}
		\caption{STO: Training the DNN with validation set.}
		\label{stowv}
	\end{subfigure}\\
	\begin{subfigure}[t]{0.95\linewidth}
		\centering\includegraphics[width=\linewidth]{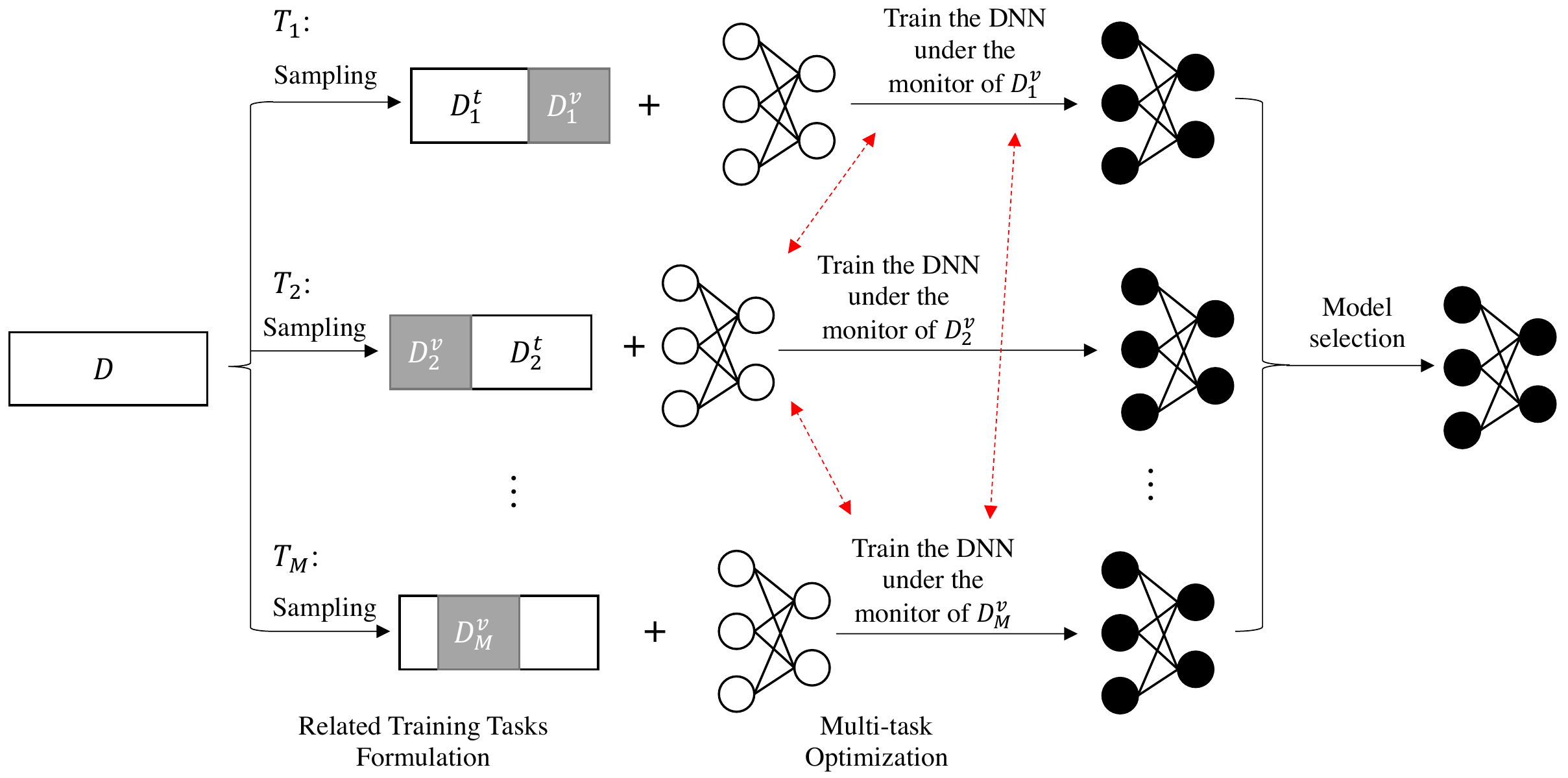}
		\caption{The proposed MTO based DNN training framework.}
		\label{mto}
	\end{subfigure}
	\caption{The existing and the proposed MTO based framework of training DNN. Each task in the proposed framework obtains a pair of training and validation sets via a certain sampling method. It uses the training set for training and validation set for monitoring the generalization ability of its trained DNNs. During the training process, the intermediate knowledge from each task is shared across all tasks to help their training where the knowledge transfer is represented by the red dashed line. At the end of the training process, the model which achieves overall best performance across all the validation sets from all pairs is selected as the final outcome.}
\end{figure*}

\subsection{Framework}
The conventional way of training DNNs corresponds to minimize a loss function which measures the mismatch between the output of the network w.r.t. an input and the actual output, which can be regarded as a single task optimization (STO) problem. One common training paradigm can be defined as follows: given a training set $D=\{x_i,y_i\}^{n}_{i=1}$ where $x_i$ and $y_i$ refer to the input and the corresponding actual output, it trains a DNN $f(;\theta)$ until reaching the pre-specified maximum epochs (\figurename~\ref{stowov}) to minimize a particular loss function \eqref{loss},
\begin{equation}
    \min_{\theta}J(\theta|D)
    \label{loss}
\end{equation}
where we define $J(\theta|D) = \frac{1}{n}\sum_{i=1}^{n}J(f(x_i;\theta), y_i)$. Once the training is completed, the trained DNN is evaluated on the testing set $D'=\{x'_j,y'_j\}^k_{j=1}$ which is invisible during training. Another popular training paradigm partitions the gross training set into one training set $D^t$ and one validation set $D^v$ via a certain sampling method and use the $D^v$ to estimate the generalization performance (evaluated by the validation loss $J(\theta|D^v)$) of the trained DNN during the training process (\figurename~\ref{stowv}). The training process will be terminated if the validation loss cannot be much reduced and the trained DNN with minimum validation loss is regarded as the final trained model. 


Unlike the above conventional training paradigms, our proposed training framework has two modules which are \textit{related training tasks formulation} and \textit{MTO} (\figurename~\ref{mto}). In the first module, we would formulate multiple related training tasks $\{T_m\}_{m=1}^M$, where each task obtains a distinct pair of training and validation sets $\{D^t_m,D^v_m\}_{m=1}^M$ via a certain sampling method. Then each task is to use the $D^t_m$ to train one individual DNN model with a pre-specified network structure and $D^v_m$ for monitoring the generalization performance during the training process. 

After that, the MTO module will solve all tasks simultaneously, and apply knowledge transfer across all tasks to help them find better model parameters which produce a lower validation loss on their associated validation set during the training process. At last, one trained DNN which achieves the overall best performance across all the validation sets from all pairs will be selected as the final trained model. The conventional STO based training method (with validation set, \figurename~\ref{stowv}) can be regarded as a special case of our proposed framework when there is only a single training task.

In this framework, the training set in each task is different and accordingly, the model learned in each task may contain the useful knowledge (e.g. promising parameter values) which can be transferred and shared with some other tasks to help their training processes to escape from inferior local optima. Meanwhile, the validation sets in different tasks provide the estimate of generalization from different perspectives. As a result, knowledge transfer and sharing across different tasks may impose implicit regularization on the training process of one task from other training processes of other tasks, aiming to produce a DNN with improved generalization which can perform well on all validation sets.

\subsection{Implementation}

\subsubsection{Formulating Multiple Related Training Tasks}
In this implementation, we formulate each training task like this: firstly, we randomly split a ratio of samples from the gross training set as validation set and the remaining as training set to form the pair $\{D^t_m,D^v_m\}$; secondly, we use the pair to formulate a training tasks $T_m$ \eqref{task} which aims to train one individual DNN model with a pre-specified network structure via its pair of training and validation sets.
\begin{equation}
    T_m:~\min_{\theta_m}J(\theta_m|D_m^t)
    \label{task}
\end{equation}
During training, the $D^v_m$ is used to estimate the change of the generalization ability of the $\theta_m$ and also provides a way to evaluate if the knowledge from other tasks is beneficial to improve the generalization ability of task $T_m$.

This process is repeated $M$ times to formulate $M$ training tasks $\{T_m\}_{m=1}^M$. These tasks are highly related since they are sampled from the same gross training set.

\subsubsection{Adaptive Multi-Task Optimization based DNN Training Algorithm}
We propose an adaptive MTO based DNN training algorithm (AMTO) which targets at solving all tasks simultaneously and transfer the intermediate learned knowledge (which we define it as model parameters $\theta$) across all tasks to improve their training performance. To effectively transfer knowledge across tasks, especially when a large number of tasks are solved together, the formulated training tasks can (i) learn task relationship with other tasks so that knowledge transfer is more likely to occur between related tasks, and (ii) determine whether to accept the transferred knowledge based on whether it can help it improve on generalization performance during the training process.


Specifically, each task maintains a relationship list ($RL$) which records how it is related to other tasks. For the $T_m$, its relationship list $RL_m$ is represented as \eqref{rl},
\begin{equation}
    RL_m=[r_1,r_2,\dots,r_M]
    \label{rl}
\end{equation}
where the $r_j\in[-\infty,+\infty]$ represents the degree of relationship to $T_j$. Then we convert the elements of $RL_m$ to the probabilities of acquiring knowledge from the corresponding tasks which sum to one by softmax function \eqref{softmax}.
\begin{equation}
    p_j=\frac{e^{r_j}}{\sum^M_{k=1,k\neq m}e^{r_k}}, j\neq m
    \label{softmax}
\end{equation}
Apparently, a higher $r_j$ represents a higher possibility of acquiring knowledge from $T_j$.

At the beginning of the algorithm, all elements of $RL$ are initialized as zero. Then for the $T_m$, it selects $T_j$ at random with the probability generated from \eqref{softmax} and acquires the model parameters $\theta_j$ as $\overline{\theta}_m$. We name this operation as \textit{knowledge reallocation}.
\begin{equation}
T_m': \min_{\overline{\theta}_m}J(f(D_m^t;\overline{\theta}_m)),~\overline{\theta}_m\gets\theta_j
\label{task_slave}
\end{equation} 
After that, a temporary DNN training task $T_m'$ is formulated as \eqref{task_slave} to evaluate if training the $\overline{\theta}_m$ on $D_m^t$ can achieve better generalization estimated on $D_m^v$ than $\theta_m$. Then the model parameters of all tasks, including $\{T_m\}_{m=1}^M$ and $\{T_m'\}_{m=1}^M$, are trained for $c$ iterations simultaneously via a gradient descent based method. Next, for each task $T_m$, we evaluate the validation loss $J(\theta_m|D_m^v)$ and $J(\overline{\theta}_m|D^v_m)$ on their corresponding validation sets and substitute the $\theta_m$ with $\overline{\theta}_m$ if the latter one achieves lower validation loss. In this way, the task $T_m$ can actively accept the knowledge from other tasks if that knowledge helps to improve its generalization performance and decline it otherwise. We name this operation as \textit{determining transfer}. Meanwhile, the relationship list will be updated. Specifically, the $r_j$ will be updated by \eqref{update}.
\begin{equation}
    r_j \gets r_j + tanh(J(\theta_m|D^v_m)-J(\overline{\theta}_m|D^v_m))
    \label{update}
\end{equation}
In other words, the $r_j$ increases if the transferred $\theta_j~(\overline{\theta}_m)$ after training on $D^t_m$ can achieve lower validation loss on $D^v_m$ than $\theta_m$ and vice versa. After this operation, the algorithm goes back to operation \textit{knowledge reallocation} or terminates if it reaches the maximum training iterations or the validation loss of any tasks does not reduce for $p$ consecutive validations.

Through periodically implementing the \textit{knowledge reallocation} and \textit{determining transfer}, each task can share its learned knowledge with other tasks and can investigate if the knowledge from other tasks is beneficial to improve its own generalization performance. In this process, once a task gets trapped into inferior local optima (i.e., unable to further reduce the validation loss), the knowledge from other tasks can potentially help it escape there. Meanwhile, the transferred knowledge from different training tasks imposes implicit regularization on the trained DNNs, which improves on the generalization performance. At the end of the training process, the DNN which achieves the highest harmonic accuracy ($A_{har}$) across all the validation sets $\{D_m^v\}_{m=1}^M$ is selected as the final output. Equation \ref{har} shows the $A_{har}$ where the $A_m$ represents the accuracy evaluated on validation set $D_m^v$.  
\begin{figure}[bp]
	\centering\includegraphics[width=\linewidth]{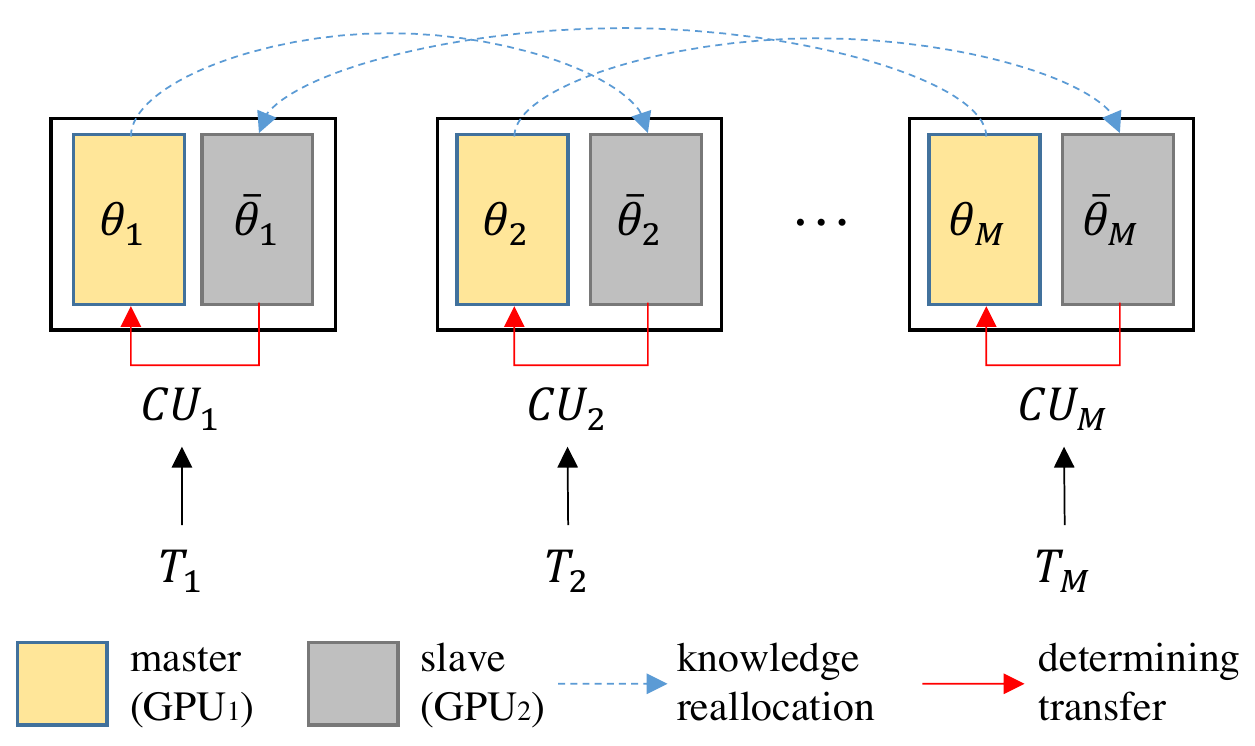}
	\caption{Computing architecture and the knowledge transfer process.}
	\label{arch}
\end{figure}

\begin{equation}
    A_{har} = \frac{M}{\sum_{m=1}^M 1/A_m}
    \label{har}
\end{equation}

\begin{algorithm}[htbp]
	\caption{Parallelized Adaptive Multitask Optimization based DNN Training Algorithm}
	\hspace*{\algorithmicindent} \textbf{Input}: $D_{train}$: training set,
	$J$: loss function, $MaxIter$: maximum training iterations, $c$: number of training steps in the interval of two consecutive \textit{knowledge allocation} operation.\\
	\hspace*{\algorithmicindent} \textbf{Output}: Trained DNN parameters $\theta$.
	\begin{algorithmic}[1]
		\ParFor {$m = 0 \to M$}     \Comment{initialization}
		\State Generate a pair of training and validation sets $\{D^t_m,D^v_m\}$ from the $D_{train}$ via random splitting.
		\State Formulate task $T_m$ which trains the DNN on $D^t_m$ and uses $D^v_m$ for estimating the generalization performance.
		\State Assign task $T_m$ to the \textit{master} of $CU_m$.
		\State Copy the task $T_m$ to the \textit{slave} of $CU_m$ and name its model parameters as $\overline{\theta}_m$.
		\State Initialize all elements of $RL_m$ as zero.
		\EndParFor
		\State $Iter \gets 1$.
		\While {$Iter*c < MaxIter$ or not trigger early stopping}
		\ParFor {$m = 0 \to M$}
		\State Acquire $\theta_j$ according to \eqref{softmax}.
		\State $\overline{\theta}_m \gets \theta_j, m\neq j$.  \Comment{knowledge reallocation}
		\State Train both $\theta_m$ and $\overline{\theta}_m$ on $D_m^t$ for $c$ iterations.
		\If{$J(\overline{\theta}_m|D_m^v)<J(\theta_m|D_m^v)$} 
		\State $\theta_m \gets \overline{\theta}_m$.		\Comment{determining transfer}
		\EndIf
		\State Update $r_j$ by \eqref{update}.
		\EndParFor
		\State $Iter \gets Iter+1$.
		\EndWhile
		\State Select one model from $\{\theta_m\}_{m=1}^M$ which achieves the highest harmonic accuracy across all validation sets $\{D_m^v\}_{m=1}^M$ as the final learned outcome.
	\end{algorithmic}
	\label{algo}
\end{algorithm}

\subsubsection{Parallelization of the implementation on a GPU cluster}
The algorithm is well-suited for parallelization to improve efficiency. We implement the algorithm on a GPU-enabled supercomputer called OzSTAR\footnote{https://supercomputing.swin.edu.au/}. As demonstrated in \figurename~\ref{arch}, we define the \textit{Computing Unity} ($CU$) as a basic unit to solve one training task $T_m$ which consists of two GPUs. These two GPUs act as \textit{master} and \textit{slave} respectively, where the \textit{slave} solves the temporary training task $T_m'$ and the \textit{master} solves the $T_m$. During training, all formulated training tasks are solved simultaneously where each $CU$ solves one task. In this case, the efficiency is comparable to the conventional STO based training paradigm. We present the pseudocode of the parallelized AMTO in Algorithm \ref{algo}.

\section{Experiments\label{ex}}

This section evaluates the proposed AMTO method on three publicly available image classification datasets, aiming to demonstrate, 
\begin{itemize}
	\item The proposed AMTO algorithm can achieve better generalization performance than the conventional STO.
	\item The performance of the proposed AMTO algorithm improves with the number of formulated tasks.
\end{itemize}
We will elaborate on the datasets details, experimental setting, and results with analysis.

\subsection{Datasets}
\paragraph{UCMerced} This dataset \cite{yang2010bag} was manually extracted from the USGS National Map Urban Area Imagery collection for various urban areas around the country. The pixel resolution of this public domain imagery is 0.3m. This dataset contains 21 land-use classes with 100 images per class, and each image is in a size of $256 \times 256$ pixels. We randomly divide this dataset into a gross training set (80\%) and a testing set (20\%).

\paragraph{OxfordPets} This dataset \cite{parkhi2012cats} has around 7400 images, which contains 37 different breeds of pets. This dataset has been pre-partitioned into a gross training set (50\%) and a testing set (50\%). The relatively small ratio of the training set increases the challenge of training a DNN with good generalization ability. 

\paragraph{RSSCN7} This dataset \cite{zou2015deep} contains 2800 remote sensing images which are from 7 typical land-use classes. There are 400 images per class collected from Google Earth which are sampled on 4 different scales with 100 images per scale. Each image has a size of $400 \times 400$ pixels. This dataset is rather challenging due to the wide diversity of the scene images which are captured under different seasons and various weathers, and sampled with different scales. Same as UCMerced, this dataset is randomly partitioned into a gross training set (80\%) and a testing set (20\%).

We will further generate training set and validation set based on the gross training set for training and use the testing set for testing.

\begin{table}[bp]
  \centering
  \caption{Comparison of the mean Top-1 accuracy(\%) of 5 runs of STO and the AMTO on the testing set of three datasets and three different DNN models.}
    \begin{tabular}{|l|r|r|r|}
    \hline
    \multicolumn{4}{|c|}{UCMerced} \\
    \hline
    Method & \multicolumn{1}{l|}{SqueezeNet} & \multicolumn{1}{l|}{MobileNetV2} & \multicolumn{1}{l|}{DenseNet-121} \\
    \hline
    STO   & 93.48 & 97.24 & 97.76 \\
    \hline
    \textbf{AMTO} & \textbf{94.95} & \textbf{97.29} & \textbf{98.05} \\
    \hline
    \multicolumn{1}{|r|}{(\textit{gap})} & 1.47 & 0.05 & 0.29 \\
    \hline\hline
    \multicolumn{4}{|c|}{OxfordPets} \\
    \hline
    Method & \multicolumn{1}{l|}{SqueezeNet} & \multicolumn{1}{l|}{MobileNetV2} & \multicolumn{1}{l|}{DenseNet-121} \\
    \hline
    STO   & 82.90  & 90.65 & 93.00 \\
    \hline
    \textbf{AMTO} & \textbf{84.34} & \textbf{91.34} & \textbf{93.19} \\
    \hline
    \multicolumn{1}{|r|}{(\textit{gap})} & 1.44 & 0.69 & 0.19 \\
    \hline\hline
    \multicolumn{4}{|c|}{RSSCN7} \\
    \hline
    Method & \multicolumn{1}{l|}{SqueezeNet} & \multicolumn{1}{l|}{MobileNetV2} & \multicolumn{1}{l|}{DenseNet-121} \\
    \hline
    STO   & 93.71 & 96.04 & 96.79 \\
    \hline
    \textbf{AMTO} & \textbf{94.75} & \textbf{96.43} & \textbf{96.89} \\
    \hline
    \multicolumn{1}{|r|}{(\textit{gap})} & 1.04 & 0.39 & 0.1 \\
    \hline
    \end{tabular}%
  \label{tab:result}%
\end{table}%

\begin{figure*}[htbp]
	\centering
	\includegraphics[width=\linewidth]{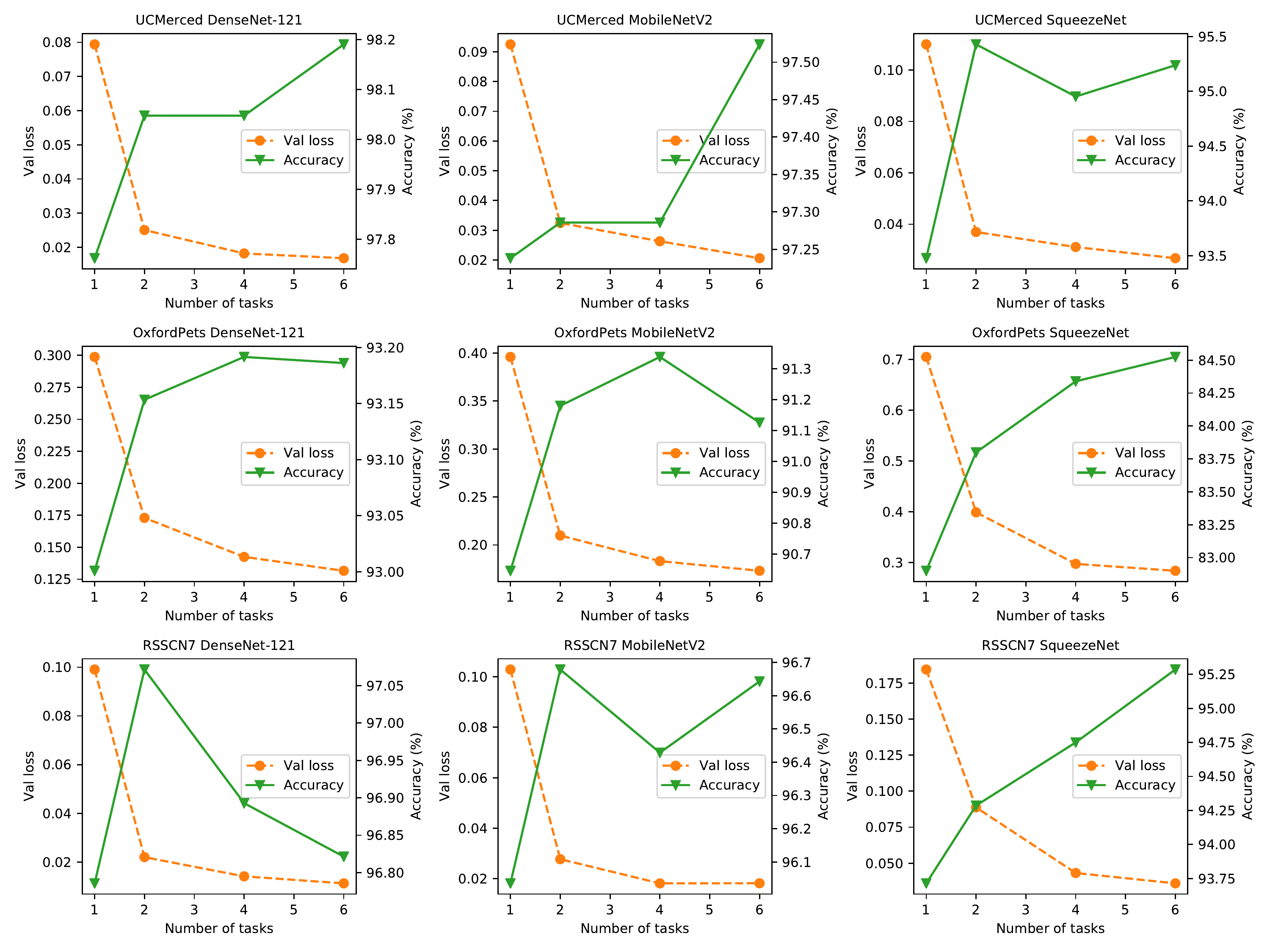}
	\caption{The mean validation loss and Top-1 accuracy of 5 runs with the different number of formulated tasks of AMTO.}
	\label{scale}
\end{figure*}

\subsection{Experimental Setting}

We compare our method with the conventional STO training paradigm (training one individual DNN with validation set) on various popular DNN models, including SqueezeNet \cite{iandola2016squeezenet}, MobileNetV2 \cite{sandler2018mobilenetv2}, and DenseNet-121 \cite{huang2017densely}. In the STO, we randomly split 10\% data for validation and the remaining for training from the gross training set of each dataset. In the MTO, we formulate each related training task with distinct pair of validation set and training set generated from the gross training set where the ratio for validation set is 10\% as well. 

Since the datasets we use are small, we initialize these DNNs via the parameters pre-trained on ImageNet \cite{deng2009imagenet}. The training samples are augmented by random horizontal flipping and resized to ($224\times224\times3$) pixels to match the required input size of the DNN models. Each single task is solved by momentum SGD with Nesterov \cite{nesterov1983method} where the initial learning rate is $1e^{-3}$ and the momentum is 0.9. The maximum training iterations are $1e^4$ and the mini-batch size is 64. The learning rate is dropped by 0.1 in $2e^3, 7e^3$ iterations. We apply early stopping both on the STO and AMTO and the training process will be terminated if the validation loss of any tasks does not reduce after 10 consecutive validations. For the AMTO method, we formulate four training tasks and apply \textit{knowledge reallocation} and \textit{determining transfer} every 100 training iterations. 

The experiments are executed for 5 runs on each dataset and DNN model where the STO and AMTO start from the same random seed in one run. We use the mean Top-1 accuracy on the testing set of the 5 runs as a metric to measure the generalization performance of the trained DNN. All the experiments are implemented with Pytorch and run on the HPC platform Ozstar where each node has two Nvidia Tesla P100 GPUs.

\subsection{Results and Analysis}
\subsubsection{Comparison With the Single Task Optimization}

Training a DNN for image classification aims to find a DNN model with desirable generalization performance, which is measured by the testing performance after training. In this experiment, we compare our proposed AMTO method with the conventional STO in terms of the Top-1 accuracy to verify the effectiveness in improving generalization ability.

Table \ref{tab:result} compares the mean Top-1 accuracy of 5 runs achieved by STO and AMTO of three datasets obtained by three popular architectures. From the table we can make the following observations: (i) the DNNs trained by AMTO performs better than those of STO on all cases, which demonstrates that AMTO is effective in training the DNN with better generalization performance. (ii) The networks with a smaller capacity (SqueezeNet and MobileNetV2) generally benefit more from AMTO. This is noteworthy as the small network often performs less desirable due to the trade-off with speed and size. Improving the performance of small networks can greatly enhance their applicability, e.g., on portable devices.

\subsubsection{AMTO With Different Number of Formulated Tasks}

The prior experiments studied AMTO of four formulated tasks. We next investigate how AMTO scales with different numbers of formulated training tasks. \figurename~\ref{scale} shows the AMTO's mean validation loss, as well as the mean Top-1 accuracy on three datasets with 1,2,4,6 formulated tasks, where one formulated task represents the conventional STO. From this figure, we can find that the mean validation loss of the target task reduces as the number of formulated tasks increases in the AMTO in all cases. This demonstrates that the optimization ability of AMTO is enhanced as the number of formulated tasks increases.

On the other hand, the mean Top-1 accuracy on the testing set of the trained DNN is higher than that of STO in all cases which verifies the effectiveness in improving the DNN's generalization performance. It's also noticeable that the mean Top-1 accuracy does not monotonically improve as the mean validation loss decreases. This phenomenon is reasonable since the distribution gap exists between the validation sets and testing set so that decreasing validation loss does not guarantee improving testing performance. The other possible reason is, randomness exists in the algorithm, especially in the step of \textit{knowledge reallocation}, which causes the fluctuation. Moreover, as the total number of training iterations is fixed, an increasing number of formulated tasks may lead to less chance of transferring useful knowledge across the tasks. To further improve the stability of AMTO, a more effective way of knowledge transfer method needs to be proposed.

\section{Conclusions and Future Work\label{cl}}
We proposed a novel DNN training framework based on MTO which can not only enhance training effectiveness but also improve on the generalization performance of the trained DNN model via knowledge transfer and sharing. We implemented the proposed framework, parallelized the implementation on a GPU cluster, and applied it to three popular DNN models. Performance evaluation and comparison demonstrated that the DNN models trained via the proposed framework achieved better generalization performance than the conventional training paradigm. In the future, we plan to explore more different ways to formulate the related training tasks. Furthermore, we will perform an in-depth study on how the number of formulated training tasks influences the performance so as to devise a way to make best use of multiple related training tasks. Moreover, we plan to further enhance the modules of related training tasks formulation and MTO in the proposed framework based on some of our previous works \cite{qin2005,gong2016,feng2019}.

\section*{Acknowledgment}
This work was performed on the OzSTAR national facility at Swinburne University of Technology. The OzSTAR program receives funding in part from the Astronomy National Collaborative Research Infrastructure Strategy (NCRIS) allocation provided by the Australian Government. This work was supported in part by the Australian Research Council (ARC) under Grant No. LP170100416, LP180100114 and DP200102611, the Research Grants Council of the Hong Kong SAR under Project CityU11202418, and the China Scholarship
Council (CSC).

\bibliographystyle{IEEEtran}
\bibliography{ijcnn}

\end{document}